\def\year{2019}\relax
\documentclass[letterpaper]{article} 
\usepackage{aaai19}  
\usepackage{times}   
\usepackage{helvet}   
\usepackage{courier}  
\usepackage{url}       
\usepackage{graphicx}  
\usepackage{natbib}
\usepackage{enumitem}
\usepackage{amsmath}
\usepackage{float}
\usepackage{subfig}
\usepackage{subfloat}
\usepackage{amssymb,amsmath,amsthm}
\usepackage{multicol}
\usepackage{bbm}
\usepackage{microtype}
\usepackage{booktabs} 
\usepackage{algorithm}
\usepackage{algorithmic}
\usepackage{xcolor}
\usepackage{multirow}

\newcommand{\R}{\mathbb{R}}

\newcommand{\w}{\mathbf{w}}
\newcommand{\x}{\mathbf{x}}

\begin{document}
%
\title{Accumulating Knowledge for Lifelong Online Learning}
\author{
Changjian Shui, Ihsen Hedhli, Christian Gagn\'e\\
\texttt{\{changjian.shui.1,ihsen.hedhli.1 \}@ulaval.ca},\texttt{\{christian.gagne@gel.ulaval.ca}\\
Electrical and Computer Engineering Department, Universit\'e Laval, Qu\'ebec (Qu\'ebec), Canada \\
}  

\maketitle
\begin{abstract}
Lifelong learning can be viewed as a continuous transfer learning procedure over consecutive tasks, where learning a given task depends on accumulated knowledge --- the so-called \emph{knowledge base}. Most published work on lifelong learning makes a batch processing of each task, implying that a data collection step is required beforehand. We are proposing a new framework, lifelong online learning, in which the learning procedure for each task is interactive. This is done through a computationally efficient algorithm where the predicted result for a given task is made by combining two intermediate predictions: by using only the information from the current task and by relying on the accumulated knowledge. In this work, two challenges are tackled: making no assumption on the task generation distribution, and processing with a possibly unknown number of instances for each task. We are providing a theoretical analysis of this algorithm, with a cumulative error upper bound for each task. We find that under some mild conditions, the algorithm can still benefit from a small cumulative error even when facing few interactions. Moreover, we provide experimental results on both synthetic and real datasets that validate the correct behaviour and practical usefulness of the proposed algorithm. 
\end{abstract}

\section{Introduction}
In the general framework of machine learning, the learning procedure can be viewed as a system (referred to as \emph{agent} in this paper) that runs an algorithm on a given dataset in order to return a hypothesis for predicting the unseen data. Typically, these algorithms require a large amount of data in order to make predictions with an acceptable level of performance. \citet{chen2016lifelong} termed such a paradigm as \emph{isolated learning}, since it does not consider any other related information or previously learned knowledge. Instead, humans have the ability to continually learn over time by accommodating new knowledge while retaining previously learned experiences. Such a continuous learning procedure has represented a challenging problem for machine learning and for the development of artificial intelligence.

This continual learning idea has inspired various machine learning strategies. In \emph{domain adaptation} \citep{pan2010survey}, the goal is to transfer the knowledge from a given task (i.e., source domain) to another task with few labeled or unlabeled observations (i.e., target domain). In the case of \emph{multi-task learning} \citep{zhang2017survey}, the goal is to set a good performance on different but related tasks simultaneously. Concerning \emph{lifelong learning} \citep{mitchell1993explanation,chen2016lifelong}, the learning procedure can be viewed as a continuous transfer learning procedure over incrementally available tasks from the underlying distribution. 
\begin{figure}[t]
\centering 
	\includegraphics[width=0.5\textwidth]{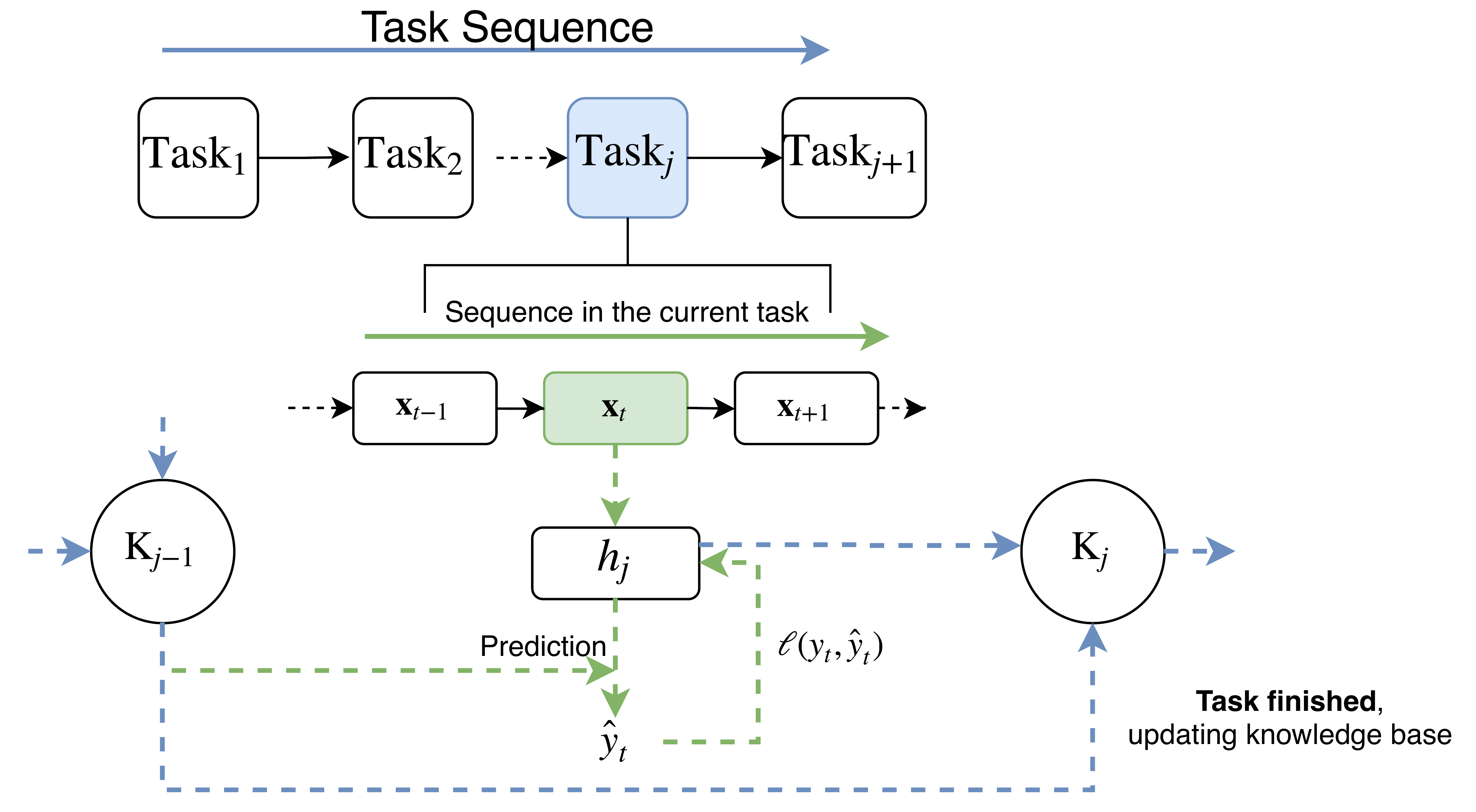}
	\caption{Two levels of sequences in lifelong online learning: 1) task sequence (blue arrows) and 2) instance sequence in each task (green arrows). For task $j$ at time $t$, the instances $\x_t$ also arrive sequentially, while in the general lifelong learning, each task is processed in a batch manner.}
\label{fig:online_lifelong}
\end{figure}

Lifelong learning has recently received increasing attention due to its implications in autonomous learning agents and robots. However, in most published work of lifelong learning, e.g., \citep{mitchell1993explanation,ruvolo2013ella,pentina2014pac}, the learning procedure for each task is \emph{offline}. That means the algorithm is generally based on the Empirical Risk Minimization (ERM) rule to obtain a good generalization performance for the unseen tasks from the same distribution. Nonetheless, there exist considerable real scenarios in which the learning procedure for each individual task is \emph{interactive}. For example, in the personalized product recommendation system, we suppose that a \emph{knowledge base} from current clients has been constructed, which can predict the preferences of existing clients. For an unknown new client, current lifelong learning algorithms must first collect a batch of data, then train such a batch with the help of the knowledge base, which cannot effectively handle the interaction demands. From this point of view, we should consider a new paradigm -- \emph{lifelong online learning} in which the agent interacts with users in each task, as showed in Figure~\ref{fig:online_lifelong}.

As opposed to the usual theoretical settings of lifelong learning, in lifelong online learning we do not assume the task generation distribution. Then the tasks are not necessarily generated in an i.i.d. fashion but are rather arbitrarily or even adversary generated, which makes this problem more challenging. Moreover, in the batch lifelong learning framework, only the total number of tasks may be unknown to us such as in \citep{ruvolo2013ella}. While in the lifelong online learning, both the total number of tasks and the number of instances in each task can be unknown for us because both of them arrive sequentially.

Motivated by the practical and theoretical considerations, we develop a new algorithm in which the agent can interactively learn with the observation in each task and also benefit from the contextual information learned by the accumulated knowledge.  At each task, the algorithm will predict the arriving data via a combination of two aspects: the predictions provided by the accumulating knowledge and the current classifier constructed by the agent from scratch in the current task. As the interaction times increase, the prediction by the current classifier will gradually play a more dominant role, because the agent has continuously learned from the instances and will be more confident about the current task. This mechanism will be useful when facing a long term interactions and overcoming the possible \emph{negative transfer} from the knowledge base \citep{pan2010survey}. 

The contribution of this paper can be summarized as follows:
\begin{itemize}
\item  We propose a computationally efficient algorithm in the lifelong online classification, which effectively leverages the information from the accumulated knowledge and the classifier which we have constructed for the current task.
\item  We theoretically provide an upper bound of the cumulative error of the proposed algorithm when facing a new task. We find that under some mild conditions, our approach can still benefit a small cumulative error even in the non i.i.d. task generation distribution.
\item Our empirical results on both synthetic and various real datasets show good performances compared with some baseline approaches.
\end{itemize}

The rest of the paper will be organized as follows. We first introduced the related work in this field. Secondly we set up the problem and discuss the proposed algorithm. In the third part, we derive the theoretical bound for the algorithm. Finally the experimental results validate the proposed algorithm.

\section{Related works}

\paragraph{Batch Lifelong learning} From the practical point of view, the lifelong learning approaches can be categorized as two families: \emph{parameter transfer} and \emph{representation transfer} \citep{chen2016lifelong}. In parameter transfer, the agent uses previously learned model parameters to help learn model parameter in the current task \citep{evgeniou2004regularized,fei2016learning}. As for the representation transfer, each task shares a common representation in a lower dimensional subspace and generally the \emph{sparse dictionary learning} approach is developed, e.g., \citep{ruvolo2013ella,maurer2013sparse,sun2018active}. While in the theoretical understanding of lifelong learning, \citet{baxter2000model} proposed the generalization bound on the VC dimension, while \citet{pentina2014pac} analyzed the lifelong learning through PAC-Bayesian theory. As for the the non i.i.d. task generation assumption, \citet{pentina2015lifelong} analyzed lifelong learning with the task environment changing over time but limited in a consistent manner. \citet{pentina2016lifelong} proposed a weighted majority vote algorithm to theoretically prove the sample complexity reduction phenomena in the \emph{lifelong learning}. 

\paragraph{Online transfer/multi-task/lifelong learning} Online transfer learning and online multi-task can also be viewed as a sequential learning procedure from some \emph{prior} knowledge. In online transfer learning, the knowledge is transferred only once from the source domain to the target domain \citep{zhao2014online,wang2015online}. In online multi-task learning \citep{saha2011online,ruvolo2013active,murugesan2017multi}, the goal is rather to learn the task relationship and try to minimize the cumulative errors on all tasks. 

There are only few published works in the lifelong online learning. For instance, \citet{alquier2017regret} proposed a similar concept of \emph{online-within-online} lifelong learning based on the representation transfer, where the data arrives sequentially in each task.  However this approach is not exactly \emph{online} since the high computational burden and storage requirements (i.e., all observed data) make it not scalable for real dataset. \citet{denevi2018incremental} proposed an improved algorithm but applied on the linear regression and adapted to the i.i.d. generation task. Moreover, the representation transfer based algorithms are still vulnerable when treating the non i.i.d. task problem. Indeed, a simple adversarial strategy is to generate instances according to two distributions which hold the same sub-feature space but different labeling distribution, for which a representation-based approach can suffer from an important cumulative error.

\section{Problem setup}

Let us define $\{j\}_{j=1}^\infty $ as an ordered set of tasks. For each task $j$, let us denote $N_{j}$ the total number of instances $\{(\x_t^{(j)},y_t^{(j)})\}_{t=1}^{N_{j}}$ where $\x_t^{(j)} \in \R^d$, $y_{t}^{(j)}\in \{-1,+1\}$, and $h_j$ as the corresponding classifiers.

Suppose that we are at task $T+1$ and the $N_{T+1}$ examples in the task $T+1$ will arrive sequentially ($N_{T+1}$ might be unknown to us), we denote $\mathrm{K}_{T}$ the current \emph{knowledge base} that contains all the historical classifiers up to the last completed task  $T$ (i.e., $K_{T}=\{h_j\}_{j=1}^{T} $).  In the beginning of the interactive learning procedure of the current task $T+1$, the classifier $h_{T+1}$ cannot make an exact decision since it has not observed a sufficient number of instances. However, $\mathrm{K}_{T}$ can provide some contextual information which can help the learner to perform better. Therefore, the final prediction rule $\mathcal{P}_{T+1}$ of the current task $T+1$ will involve two parts, as shown in the prediction model given in Figure~\ref{fig:online_lifelong}.

\section{Proposed algorithm}

The proposed lifelong online algorithm proceeds by using the contextual information available in the accumulated knowledge to perform better in the current learning procedure. For the current task $T+1$, two prediction stages are performed.
\begin{description}[labelwidth=0in,labelindent=0em,leftmargin=0\labelsep]
\item[Prediction from the knowledge base ($O_{1:T}$):] constructed classifiers $\{h_j\}_{j=1}^{T}$ are evaluated on the new task $T+1$ using:
\begin{equation}
    h_j(\x_t^{(T+1)}) = \langle \w^{(j)}, \x_t^{(T+1)} \rangle,
    \label{pred_rule}
\end{equation}
where $\w^{(j)}$ is the parameter corresponding task $j \in [1, T]$. This accumulated knowledge is used through the application of an \emph{expert model} to make predictions over the arriving observations through the pool of the previous models, as shown in~\citep{cesa2006prediction}.
\item[Prediction from the current classifier ($O_{T+1}$):] current model $h_{T+1}$ is being interactively updated by the arriving observations $\{\x_t\}_{t=1}^{N_{T+1}}$ and the prediction rule is the same as Equation \ref{pred_rule}.
\end{description}

Updates to the current model $h_{T+1}$ are being interactively done via \emph{online gradient descent} of the regularized convex loss~\citep{hazan2016introduction}: $$\mathcal{L}(\w)= \ell_t(\w) + \frac{\lambda}{2}\|\w\|^2.$$ In the present paper we fix $\ell_t$ to be the \emph{hinge loss}, i.e.: $\ell_t(\w^{(T+1)})~=~\max\{0,1-y_t\langle \w^{(T+1)},\x_t \rangle \},$ where $\w^{(T+1)}$ is the vector of parameters corresponding task $(T+1)$.

We propose to balance between the prediction $O_{T+1}$ using only the current task and the prediction $O_{1:T}$ using the accumulated knowledge base through a non-increasing series $\{\alpha_t\}_{t=1}^{N_{T+1}}$ with $\forall t,~0\leq\alpha_{t+1}\leq\alpha_{t}\leq 1$ as the following: $$\mathcal{P}_{T+1}(\x)=\alpha_t O_{1:T}(\x) + (1-\alpha_t) O_{T+1}(\x).$$ Intuitively, when gradually receiving more and more instances, the impact of the knowledge base on the final prediction will also decrease gradually. Therefore, the decreasing $\alpha_t$ aims at balancing predictions between the knowledge base and the current classifier $h_{T+1}$. And as for the first task, we only learn from the current classifier $h_{1}$ since $K_0=\emptyset$, i.e., $\mathcal{P}_{1}(\x) = O_1(\x).$ In this case one could see the problem as a classical online learning algorithm, with the predictions being performed by a classifier interactively updated by the arriving observations.

For all $T$, to evaluate $\mathcal{P}_{T+1}$, we benefit from the fact that the proposed updating rule for the knowledge base is additive. Algorithms~\ref{pred_T} and~\ref{select} present the approach more formally. Moreover, two distinct approaches (options) are proposed in order to make predictions from the knowledge base for Accumulated Knowledge Lifelong Online (AKLO) learning: \emph{AKLO\,Sum} and \emph{AKLO\,Sample}.
\begin{description}[labelwidth=0in,labelindent=0em,leftmargin=0\labelsep]
    \item[AKLO\,Sum] Predictions are made by a weighted vote from the models in the knowledge base:
\begin{equation}
O_{1:T}(\x_t^{(T+1)})= \mathcal{T}_{[-1,1]} \big( \sum_{i=1}^{T} p_t(i)  \langle \w^{(i)}, \x_t^{(T+1)} \rangle \big),\label{our_sum}
\end{equation}
where $\mathcal{T}_{[a, b]}$ is a piece-wise function defined as:  
\begin{equation*}
 \mathcal{T}_{[a,b]}(x)  = \left\{
        \begin{array}{ll}
            a & \quad x\leq a \\
            x & \quad  a< x<b\\
            b & \quad x \geq b
        \end{array}.
    \right.
\end{equation*}
    \item[AKLO\,Sample] Predictions are made by sampling a model $i$ from the knowledge base according to the Categorical distribution $i \sim \mathrm{Cat}(p_t)$:
\begin{equation}
O_{1:T}(\x_t^{(T+1)}) = \mathcal{T}_{[-1,1]} \big( \langle \w^{(i)}, \x_t^{(T+1)} \rangle \big),
\label{our_sample}
\end{equation} 
\end{description}
where the weight $p_t$ is defined as the $T$ simplex with $\sum_i^T p_t(i)~=~1,~\forall i,~p_t(i)\geq 0$, estimated by their historical behaviors $p_t(i) \propto \exp\{-\epsilon_t L_t(i)\}$. $L_t(i) = \sum_{k=1}^t e_k(i)$ represents the cumulative error at task $T+1$ for model $h_i$ ($1\leq i \leq T$) in the interactive learning until time $t$. It is also worth mentioning that since the tasks are not necessarily i.i.d. generated, the performance of the algorithm is measured by a small cumulative error in the task $T+1$.
\begin{center}
\begin{algorithm}[t]
		\caption{Prediction $\mathcal{P}_{T+1}$ at task $T+1$}
		\begin{algorithmic}[1] 
		\REQUIRE $\eta_t >0$, $\forall t \in \{1,\dots,N_{T+1}\}$
  		\ENSURE  $\w^{(T+1)}_{1} = \mathbf{0}$, $p_{1}= \frac{1}{T}\mathbf{1}$
  		
   		\FOR{$t= 1$~to~$N_{T+1}$}
		\STATE Observe $\x_t^{(T+1)}$
		\STATE Predict using the current classifier $h_{T}$ by computing the confidence:\\
		$O_{T+1}(\x_t^{(T+1)})=\mathcal{T}_{[-1,1]}\big(\langle \w^{(T+1)}_{t}, \x_t^{(T+1)}\rangle\big)$
		\STATE Predict using cumulative knowledge $K_T$ by applying Equation \ref{our_sum} or \ref{our_sample}, get confidence $O_{1:T}(\x_t^{(T+1)})$
		\STATE Final Prediction: 
		$$\hspace{-1em} \hat{y}_t^{(T+1)} =\mathrm{sign}[(1-\alpha_t) O_{T+1}(\x_t^{(T+1)}) + \alpha_t O_{1:T}(\x_t^{(T+1)})]$$
        
		\STATE Receive the real label $y_t^{(T+1)}$
		\STATE Update weight $p_t$ by $y_t^{(T+1)}$ with Algorithm \ref{select}
		\STATE Update $\w^{(T+1)}_{t}$: 
		\IF{$y_t^{(T+1)}\langle \w^{(T+1)}_{t}, \x_t^{(T+1)} \rangle \geq 1$}
		\STATE $\w^{(T+1)}_{t+1} = (1-\eta_t \lambda)\w^{(T+1)}_{t}$
		\ELSE
		\STATE $\w^{(T+1)}_{t+1} = (1-\eta_t \lambda)\w^{(T+1)}_{t} + \eta_t y_t^{(T+1)}\x_t^{(T+1)} $
		\ENDIF
   \ENDFOR
\end{algorithmic}
\label{pred_T}
\end{algorithm}
\end{center}

\begin{center}
\begin{algorithm}[t]
		\caption{Updating the knowledge base $\mathrm{K}_{T}$ }
		\begin{algorithmic}[1]
   		\REQUIRE  Knowledge $\mathrm{K}_{T}$, $\epsilon_t >0$, $\forall t\in \{1,\dots,N_{T+1}\}$
  		\ENSURE   $L_{1} = \mathbf{0}$	
  		\FOR{$t= 1$ to $N_{T+1}$}
			\STATE Receive label $y_t^{(T+1)}$
			\STATE Compute error $e_t$ for each model $j\in\{1,\dots,T \}$ in the knowledge base $K_T$:  
			$$ e_t(h_j) = \Big(\mathcal{T}_{[-1,1]}\big( \langle \w^{(j)},\x^{(T+1)}_{t}\rangle \big) - y_t^{(T+1)}\Big)^2 $$
			\STATE Update $L_{t+1} = L_{t} + e_t$ 
			\STATE Update $p_t(i) = \frac{\exp\{-\epsilon_t L_{t}(i)\}}{\sum_{j=1}^{T}\exp\{-\epsilon_t L_{t}(j)\}}$
	    \ENDFOR
\end{algorithmic}
\label{select}
\end{algorithm}
\end{center}
\paragraph{Computational complexity} The computational complexity of updating rule in Algorithm~\ref{select} is $\mathcal{O}(Td + T)$. Then plugging in Algorithm \ref{pred_T} we derive a global complexity of $\mathcal{O}(N_{T+1}(Td+T+d))$, which corresponds to a \emph{linear time} algorithm that is able to efficiently process high dimensionality datasets. 

\section{Theoretical analysis}

In the following, we analyze the error bound of the proposed algorithm at current task $T+1$. This is coherent with \cite{pentina2016lifelong}, where a batch non i.i.d. lifelong learning problem is analyzed. Two scenarios are discussed: a known horizon $N_{T+1}$ and a fixed but unknown horizon $N_{T+1}$.

\subsection{Known horizon $N_{T+1}$}

\paragraph{Theorem 1} Supposing $\|\x\|\leq X$ and $\|\w\|\leq R$, for any $\{\alpha_t\}_{t=1}^{N_{T+1}}$ such that $\alpha_1 = 1$ and $\forall t,~ 0 \leq \alpha_{t+1} \leq \alpha_{t} \leq 1$, we set $\lambda = \frac{X+R}{R}\sqrt{\frac{\log(N_{T+1})+1}{N_{T+1}}} $, $\eta_t = 1/(\lambda t)$, $\epsilon_t = \sqrt{\frac{\log(T)}{8\sum_{t=1}^{N_{T+1}}\alpha_t}}$ and we use Equation~\ref{our_sum} (AKLO\,Sum) as the prediction rule. If the algorithm predicts $\hat{y}_t$, then the cumulative error $E_{T+1} = \sum_{t=1}^{N_{T+1}}\mathbf{1}\{\hat{y}_t \neq y_t \} $ at task $T+1$ can be bounded by:
\begin{equation*}
\begin{split}
E_{T+1} \leq & \sum_{t=1}^{N_{T+1}} \alpha_t e_t(\w^{\star\star}) + \sum_{t=1}^{N_{T+1}}(1-\alpha_t)\ell_t(\w^{\star})\\
            & +4 \sqrt{2\log(T) \sum_{t=1}^{N_{T+1}}\alpha_t}\\
            & +\sum_{t=1}^{N_{T+1}}(1-\alpha_t) R(X+R) \sqrt{\frac{\log(N_{T+1})+1}{N_{T+1}}},
\end{split}
\end{equation*}
where:
\begin{align*}
\w^{\star} & = \underset{\|\w\|\leq R}{\min} \sum_{t=1}^{N_{T+1}} \ell_t(\w),\\
\w^{\star \star} & = \underset{\w\in\{\w^{(1)},\dots,\w^{(T)}\}}{\min} \sum_{t=1}^{N_{T+1}} e_t(\w).    
\end{align*}
We provide the complete demonstration in the supplementary material. 

If we set $\alpha_t \equiv 1$, then the error bound can be simplified as $\sum_{t=1}^{N_{T+1}} e_t(\w^{\star\star}) +4 \sqrt{2 N_{T+1}\log(T)}$, which exactly recovers to the error bound of the expert problem \citep{cesa2006prediction}. Furthermore, if we set $\alpha_t \equiv 0$, the error bound can be simplified as  $\sum_{t=1}^{N_{T+1}}\ell_t(\w^{\star}) +  R(X+R)\sqrt{N_{T+1}(\log(N_{T+1})+1)}$, which recovers to the error bound of the Follow The Adaptive Regularized Leader (FTARL) problem \citep{mcmahan2017survey}. We should also point out that the proposed algorithm is not sensible to the size of the knowledge base because of the $\log(T)$ term in the bound. Besides the direct conclusion from Theorem 1, we can also derive the Corollary 1 if the number of instances $N_{T+1}$ is small.

\paragraph{Corollary 1} For $\gamma \in(0,1)$, if $N_{T+1}$ satisfies $N_{T+1} \leq t_0$, with $t_0 = \max\{t|\alpha_t \geq \frac{K}{1+K}\}$, if we assume that $\zeta = \underset{\w\in\{\w^{(1)},\dots,\w^{(T)}\}}{\min} \sum_{t=1}^{N_{T+1}} e_t(\w) $ is non zero ($\zeta >0$) and $K \geq \max\{\frac{1+XR}{\gamma\zeta},\frac{R(X+R)}{4\gamma\sqrt{2\log(T)}} \}$, the cumulative error bound $E_{T+1}$ in Theorem 1 can be simplified as:
\begin{equation*}
E_{T+1} \leq (1+\gamma)\Big(\sum_{t=1}^{N_{T+1}} \alpha_t e_t(\w^{\star\star}) +  4 \sqrt{2\log(T) \sum_{t=1}^{N_{T+1}}\alpha_t} \Big).
\end{equation*}
The proof of Corollary 1 is also provided in the supplementary material. This corollary reveals an interesting fact: if $N_{T+1}$ is smaller than a predefined threshold, then the best model in the knowledge base $\mathrm{K}_T$ will play a dominant role in the error bound. Therefore, in the current task $T+1$ and even with small interaction times $N_{T+1}$, it is still possible to obtain a good performance although the current classifier $h_{T+1}$ is not well trained. 

\paragraph{Corollary 2} Supposing all the conditions of Theorem 1 hold and now we use Equation~\ref{our_sample} (AKLO\,Sampling) as the prediction rule. If the algorithm predicts $\hat{y}_t$, then with the probability higher than $1-\delta$, with $\forall \delta \in (0,1)$, the accumulative error $ E_{T+1}$ at task $T+1$ can be bounded by:
\begin{equation*}
\begin{split}
E_{T+1} \leq & \sum_{t=1}^{N_{T+1}} \alpha_t e_t(\w^{\star\star}) + \sum_{t=1}^{N_{T+1}}(1-\alpha_t)\ell_t(\w^{\star})\\
& +4 \sqrt{2 \log(T) \sum_{t=1}^{N_{T+1}}\alpha_t} + \sqrt{8\sum_{t=1}^{N_{T+1}}\alpha_t^2 \log(\frac{1}{\delta})}\\
& + \sum_{t=1}^{N_{T+1}}(1-\alpha_t) R(X+R) \sqrt{\frac{\log(N_{T+1})+1}{N_{T+1}}}.
\end{split}
\end{equation*}
The proof of Corollary 2 is based on Theorem 1 and Hoeffding-Azuma inequality. We also provide the demonstration in the supplementary material. 

\subsection{Unknown horizon $N_{T+1}$}

As we described in the introduction, in the lifelong online learning, both the number of tasks and the number of examples in each task can all be unknown to us. However, parameters $\eta_t$ and $\epsilon_t$ directly depend on $N_{T+1}$. In the following, we develop some strategies for setting these hyper-parameters without the knowledge of $N_{T+1}$.

For the current learner $h_{T+1}$, if we directly adjust $\lambda = \frac{X+R}{R}$, we can also derive a bound in this setting, where the whole proof procedure is similar to Theorem 1. As for the prediction from Algorithm \ref{select} in the knowledge base, we apply the \emph{double trick} for online learning  \citep{cesa2006prediction}, which divides the time interval into periods $I_m = [2^m,2^{m+1}-1]$ of length $2^m$, for $m=0,1,\dots$, until the task completes. The modified algorithm is almost the same as Algorithm \ref{select}, with the difference of a reset $L_{t} = 0$ at the beginning of each new interval $I_m$ and the use of $\epsilon_t = \sqrt{\frac{\log(T)}{8\sum_{j=1}^{2^m}\alpha_j}} $ for $t\in I_m$. Based on such a modified algorithm, we can derive the new error bound for the unknown $N_{T+1}$ at task $T+1$.

\paragraph{Theorem 2} Supposing $\|\x\|\leq X$ and $\|\w\|\leq R$, for any $\{\alpha_t\}_{t=1}^{N_{T+1}}$ with $\alpha_1 = 1$ and $\forall t,~0 \leq \alpha_{t+1} \leq \alpha_{t} \leq 1$, we set $\lambda = \frac{X+R}{R}$, $\eta_t = 1/(\lambda t)$, $\epsilon_t$ decided by the double trick and we choose Equation~\ref{our_sum} (AKLO\,Sum) as the prediction rule. If the algorithm predicts $\hat{y}_t$, then the cumulative error $E_{T+1}$ at task $T+1$ can be bounded by:
\begin{equation*}
    \begin{split}
        E_{T+1} \leq & \sum_{t=1}^{N_{T+1}} \alpha_t e_t(\w^{\star\star}) + \sum_{t=1}^{N_{T+1}}(1-\alpha_t)\ell_t(\w^{\star})\\
        & +  4 \log(N_{T+1}) \sqrt{2\log(T) \sum_{t=1}^{N_{T+1}}\alpha_t} \\
        & + R(X+R) \sum_{t=1}^{N_{T+1}}(1-\alpha_t). 
    \end{split}
\end{equation*}
The proof of Theorem 2 is provided in the supplementary material. We should point out that this error bound is worse than the original bound proposed in Theorem 1 since we do not know the $N_{T+1}$ in advance. However, in the real lifelong learning problem, the value $N_{T+1}$ is generally not too large for each task, such that $\log(N_{T+1})$ and $\sum_{t=1}^{N_{T+1}}(1-\alpha_t)$ are both small, making the learning procedure effective.

\section{Empirical evaluations}
We evaluate the empirical performance of the proposed algorithm in the online setting. We will test the proposed algorithm by two synthetic and three real datasets. Concerning real datasets, the data are already separated in different categories (i.e., tasks)  and for each task a uniform sampling without replacement is performed on the original data to keep only a portion of examples. The reason behind this is twofold: 1) in the lifelong learning context the number of examples for each task is generally not large; and 2) keeping a relative smaller size theoretically emphasizes the effectiveness of the proposed algorithm, as shown in Corollary~1. In the following a further description of the dataset used is given.  

\subsection{Dataset description}
\paragraph{Synthetic data 1 (syn1)}  For testing the behavior of the proposed algorithm particularity in the non i.i.d. assumption,  we create a set of tasks $ \{j\}_1^{50} $ generated by two different distributions, namely $\mathcal{D}_1$ and $\mathcal{D}_2$ (25 tasks for each distribution). Each task $j$ is composed by  $N_j = 100$ instances $\{(\x_t^j, y_t^j)\}_{t=1}^{100},$  where $\x_t^j \in \R^2$ and $y_t^j \in \{-1,+1\}$. For a given task $T$ from the first distribution $\mathcal{D}_1$, $\mathbf{x}_t^T $ is generated from the normal multivariate distribution $\mathbf{x}_t^T \sim \mathcal{N}(\boldsymbol{\mu}_1,\sigma_1\mathbf{I})$ with $\boldsymbol{\mu}_1=[10,10]$ and $\sigma_1 = 1$ and the labeling function $y_t^T$ is expressed as $y_t^T = \mathrm{sign}(\mathbf{a}_{T}^{\top}\mathbf{x}_t)$, where $\mathbf{a}_{T} = [-1+\epsilon_T,1+\epsilon_T]^{\top}$ is the decision function and $\epsilon_T \sim \mathcal{N}(0,10^{-3})$ is the variant part for task $T$.  Similarly, for a given task $T^{\prime}$ from the second distribution $\mathcal{D}_2$  the same strategy is applied using different parameter settings. Thus, for $\mathcal{D}_2$ we set  $\boldsymbol{\mu}_2 = [20,5]$, $\sigma_2 = 1$, $y_t^{T^{\prime}}= \mathrm{sign}(\mathbf{a}_{T^{\prime}}^{\top}\mathbf{x}_t)$ and the decision boundary $\mathbf{a}_{T^{\prime}} = [-0.25+\epsilon_{T^{\prime}},1+\epsilon_{T^{\prime}}]^{\top}$ with $\epsilon_{T^{\prime}}\sim\mathcal{N}(0,10^{-3})$. Furthermore, the agent has no access to the data structure and generation information and the tasks from these two distributions will be arbitrarily provided to the agent.  

\paragraph{Synthetic data 2 (syn2)}  With this dataset, we will test the performance of the proposed algorithm in a non-obvious situation (e.g., adversarial setting). For this, we adopt similar settings as in the syn1 with the same number of tasks and the same number of examples for each task.  We also keep the same generation technique from $\mathcal{D}_1$ and $\mathcal{D}_2$ . However, for all the tasks the observations have the same feature generation distribution (i.e., the same marginal distribution) $\mathbf{x}_t \sim \mathcal{N}(\boldsymbol{\mu},\sigma\mathbf{I})$ and a totally different labeling function. As for the first distribution $\mathcal{D}_1$, the labeling function is $y_t = \mathrm{sign}(\mathbf{a}^{\top}\mathbf{x}_t)$ and for the second the distribution $\mathcal{D}_2$, the labeling function is adversarial given by $y_t = \mathrm{sign}(-\mathbf{a}^{\top}\mathbf{x}_t)$, where  $\mathbf{a} = [-1+\epsilon_T,1+\epsilon_T]^{\top}$  and  $\epsilon_T$ is kept the same as in syn1. The non-obvious generation is common in the non i.i.d. settings. For example, in the personalized product recommendation for two groups of clients, the product might have the same features but these two groups may have totally different preferences on the same product. In the experiment, the agent still knows nothing except $(\x_t,y_t)$. 

\paragraph{Landmine detection\footnote{
\scriptsize\url{http://www.ee.duke.edu/~lcarin/LandmineData.zip}} (Landmine)} This dataset contains 29 binary classification tasks corresponding to  29 geographical regions. For each task, the goal is to detect landmines ($+1$) or clutters ($-1$). Each example contains $9$ features, we also add a bias term, resulting in $10$ features during the experiment. We randomly sample $150$ examples for each task. 

\paragraph{Spam detection\footnote{\scriptsize\url{http://ecmlpkdd2006.org/challenge.html}} (Spam)} We adopt the dataset from ECML PKDD 2006 Discovery challenge for the spam detection task of $14$ different users. Each user can be viewed as a individual task and the goal is to build a personalized mail filtering system. This task is a binary classification problem with label spam ($+1$) and non-spam ($-1$). Each example has an extremely high number of features ($\approx 1.5\times 10^5$) representing the word occurrence frequency (bag of word model). We randomly select $200$ examples for each task.

\paragraph{Shoes data\footnote{\scriptsize\url{http://vision.cs.utexas.edu/whittlesearch/}} (Shoes)} We used the shoes dataset with attributes from \citet{kovashka2012whittlesearch} and the same setting as in \citep{pentina2015curriculum}. In this experiment, we study the scenario of learning visual attributes that characterize shoes across different shoe models. We have $10$ attributes representing the different tasks in  the proposed lifelong online setting (\emph{pointy at the front, open, bright in color, covered with ornaments, shiny, high at the heel, long on the leg, formal, sporty, feminine}) which describes the shoes models.  Each attribute (i.e., task) is a binary  classification problem with $100$ examples: $(+1)$ means holding such a property and $(-1)$ means not. Each example has $990$ dimensional features from the original image. In this dataset, it is worth mentioning that some tasks are clearly related such as \emph{high heel} and \emph{shiny}, and some tasks are not, such as \emph{high heel} and \emph{sporty}.

\subsection{Methods and measurements}
In our experiments, we compare different baseline approaches to verify that the proposed algorithms can effectively learn from the accumulating knowledge during the lifelong online learning process:
\begin{description}[labelwidth=0in,labelindent=0em,leftmargin=0\labelsep]
	\item[ITOL:] Independent Task Online Learning --- for each task, we train the classifier in an online way without taking into account the accumulated experience from the previous tasks;
	\item[TOL:] Tasks Online Learning --- only one online classifier is performed by concatenating the data of all the tasks;
	\item[AKLO\,Sum:] Accumulating Knowledge for Lifelong Online learning using the sum updating rule described in Equation~\ref{our_sum} --- the prediction is the weighted sum of the learned classifiers in the \emph{knowledge base};  
	\item[AKLO\,Sample:] Accumulating Knowledge for Lifelong Online learning using the sampling updating rule described in Equation~\ref{our_sample} --- the prediction comes from a random sampling w.r.t. the estimated normalized weight; 
	\item[Unif\,Sum:] Uniform sum from the \emph{knowledge base} --- this approach is similar to the proposed algorithm with Equation~\ref{our_sum}, but the prediction is directly the average of the advice from the \emph{knowledge base}, i.e., $p_t(i) = \frac{1}{T}$;
	\item[Unif\,Sample:] Uniform sampling from the \emph{knowledge base} --- this approach is similar to the proposed algorithm with Equation~\ref{our_sample}, with the uniform sampling strategy.
\end{description}
To fix the hyper-parameters $\alpha_t$, $\epsilon_t$  and $\lambda$  of the proposed model, we adopt a simple linear time function for $\alpha_t$: $\alpha_t = 1- \frac{(t-1)}{N_{T+1}}$ and $\epsilon_t = \sqrt{\frac{\log(T)}{8\sum_{t=1}^{N_{T+1}}\alpha_t}}$ as described in Theorem 1. As for selecting the $\lambda$, we choose it as the best $\lambda$ from $[10^{-3},10^{-2},\dots,10^{3}]$ obtained from averaging the performance over ITOL. For some datasets which already have a training and testing set (such as Spam and Shoes datasets), we validate the best $\lambda$ from the training set. For the other datasets, for each task, we randomly sample a small portion and we shuffle the sampled examples. Lastly, the learning rate in Algorithm~\ref{pred_T} is set accordingly as $\eta_t = 1/(\lambda t)$. 

\subsection{Results and analysis}
 To measure the performance, we report the average cumulative errors (ACE) over the tasks in the lifelong learning: $\mathrm{ACE} = \frac{1}{T}\sum_{j=1}^{T}\frac{1}{N_{j}}\sum_{t=1}^{N_{j}}\mathbf{1}\{\hat{y}_t \neq y_t \}$.
 
 Table~\ref{error_table} presents 10 repetitions by randomly shuffling the task order and the example order in each task.
The proposed AKLO\,Sum and AKLO\,Sample methods demonstrate their effectiveness by using the accumulating knowledge compared to the baseline approaches. Particularly the proposed approaches are more efficient when using datasets where the tasks are not i.i.d. generated (i.e.,  syn1, syn2). 

It is worth mentioning that the Unif\,Sample is even worse than the independent training (ITOL) as \emph{negative transfers} are likely to occur with non i.i.d. datasets. By comparing results obtained with Unif\,Sum to AKLO\,Sum, it reveals that the proposed AKLO\,Sum is not simply leveraging the information from the previous experiences, but is able to make an online selection of related tasks when effectively combining them to make predictions, such as the \emph{shoes} dataset. AKLO\,Sum has about $10\%$ lower error rate than the Unif\,Sum since some tasks are not related and Unif\,Sum approach treats all the tasks equivalently.

Moreover, in the \emph{landmine} dataset, we find that the performance of the proposed methods are slightly higher than the results given by for instance Unif\,Sum technique compared to other dataset. A possible reason for this is that \emph{landmine} is generally regarded as an i.i.d. realization by the task distribution \cite{pentina2014pac} and Unif\,Sum behaves similarly to a \emph{parameter transfer} approach in this case.

\begin{figure*}[!tbp]
		\centering 
		\subfloat[syn1]{\includegraphics[width=0.33\textwidth]{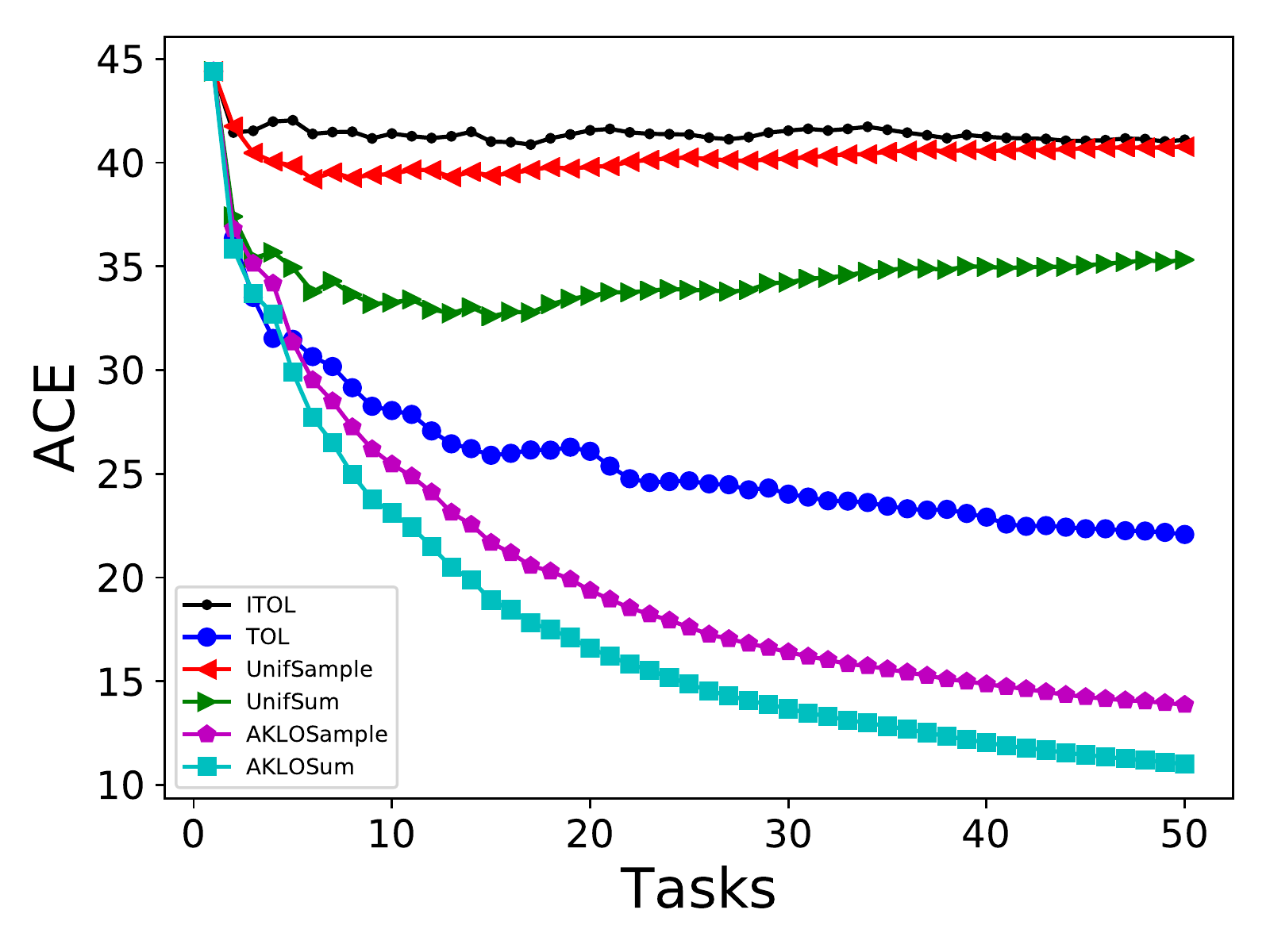}}
		\subfloat[syn2]{\includegraphics[width=0.33\textwidth]{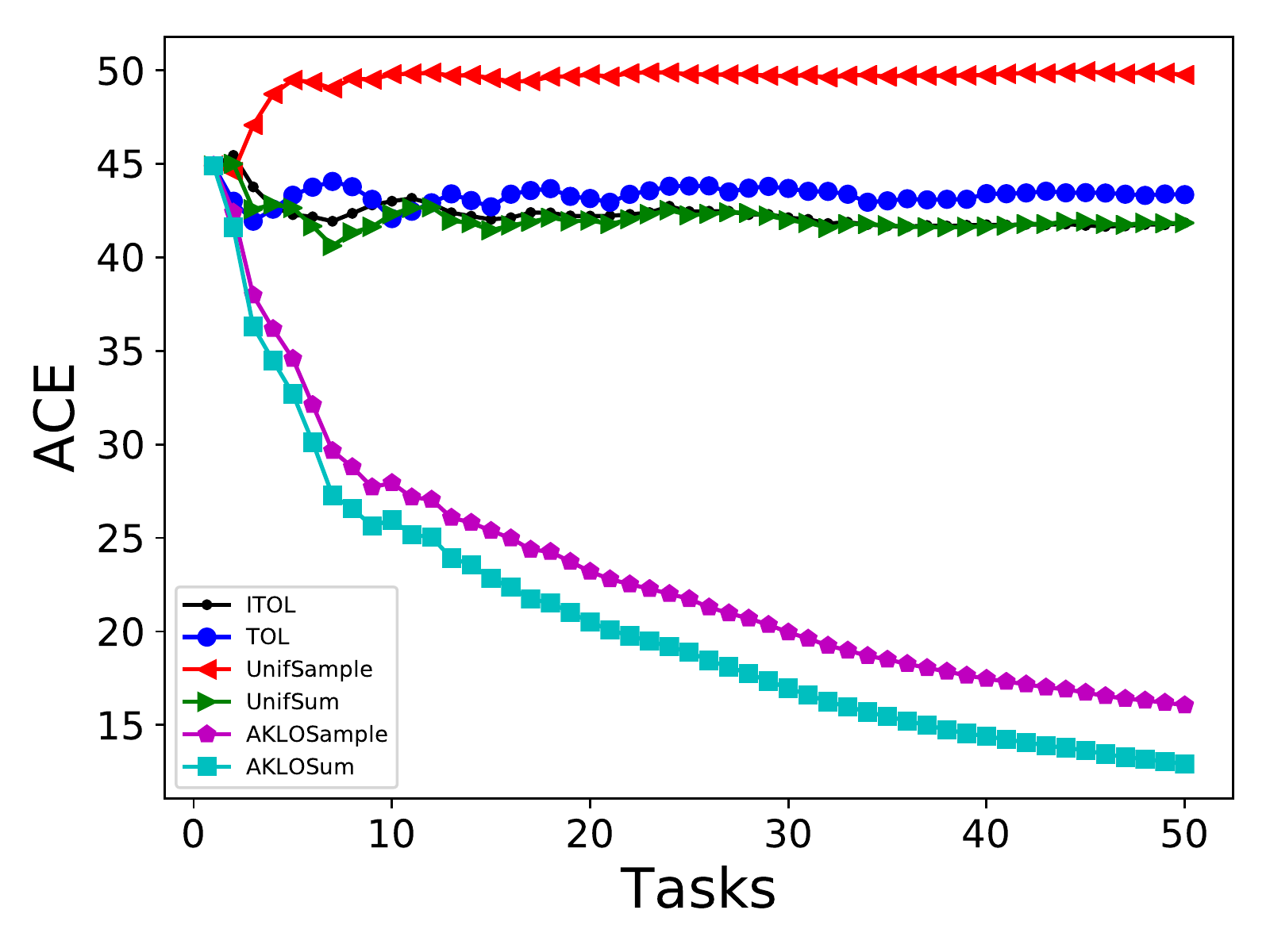}}
		\subfloat[Landmine]{\includegraphics[width=0.33\textwidth]{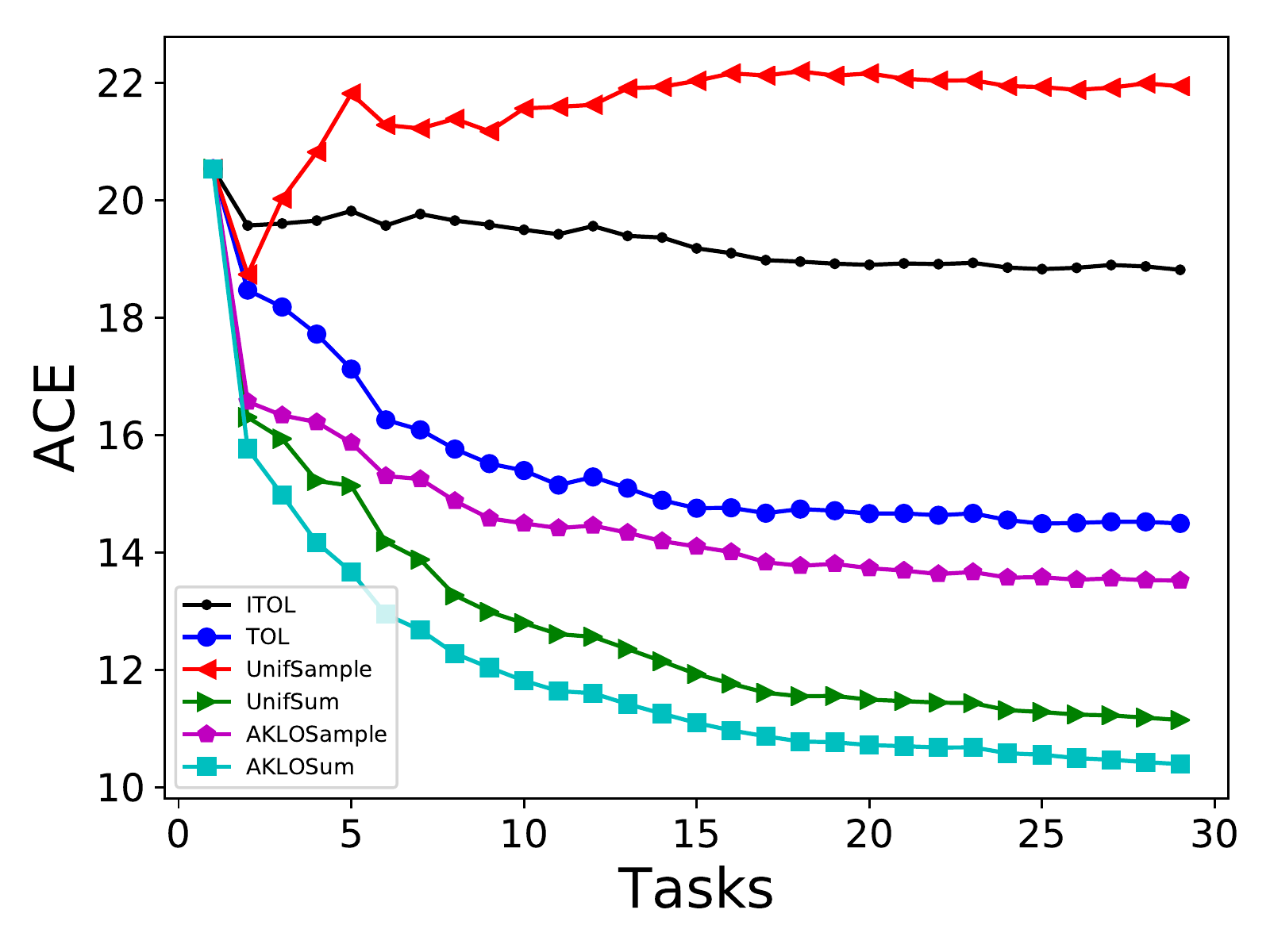}}\\
		
		\subfloat[Spam]{\includegraphics[width=0.33\textwidth]{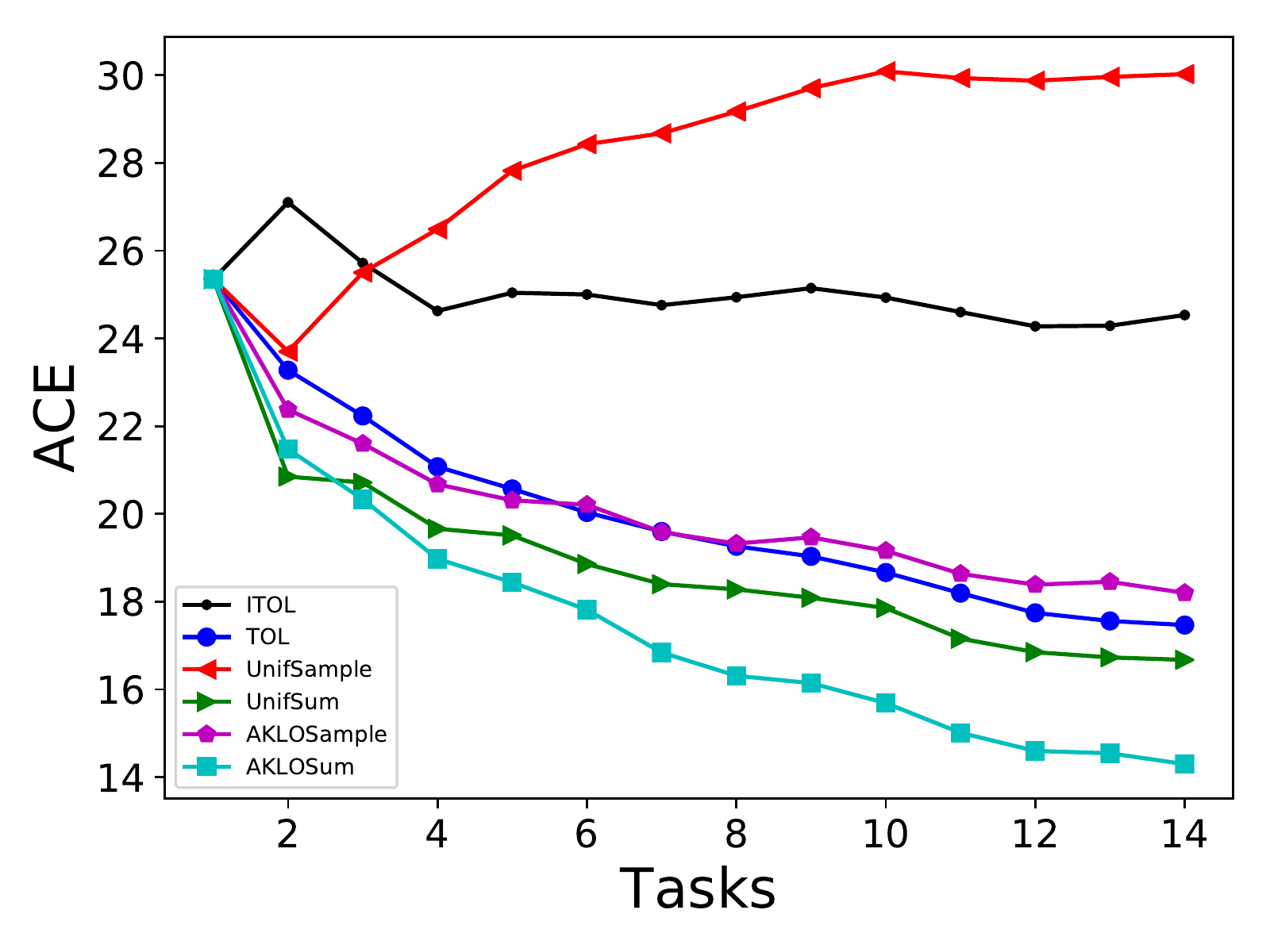}}
		\subfloat[Shoes]{\includegraphics[width=0.33\textwidth]{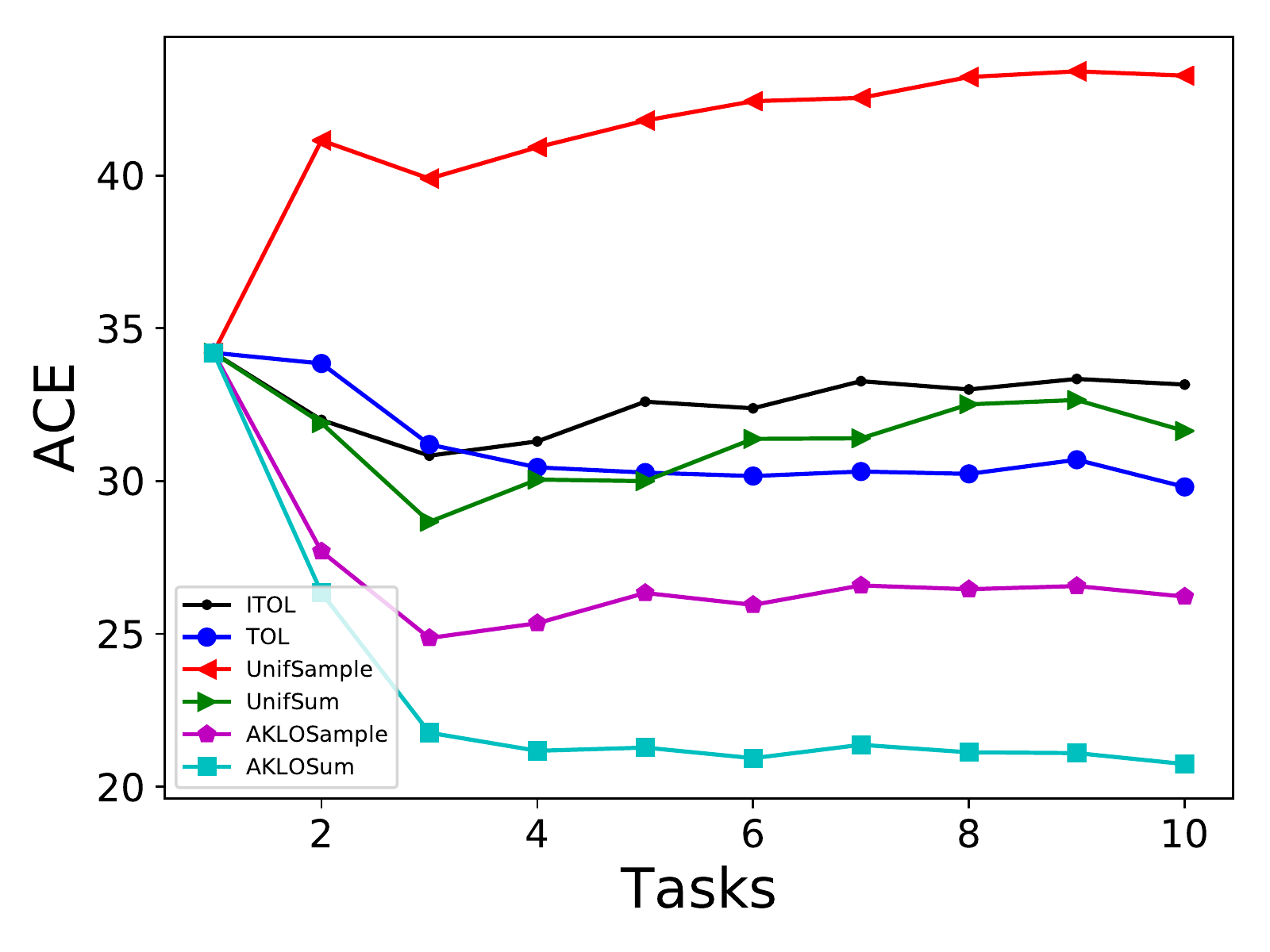}}
		\subfloat[Shoes' last task]{\includegraphics[width=0.33\textwidth]{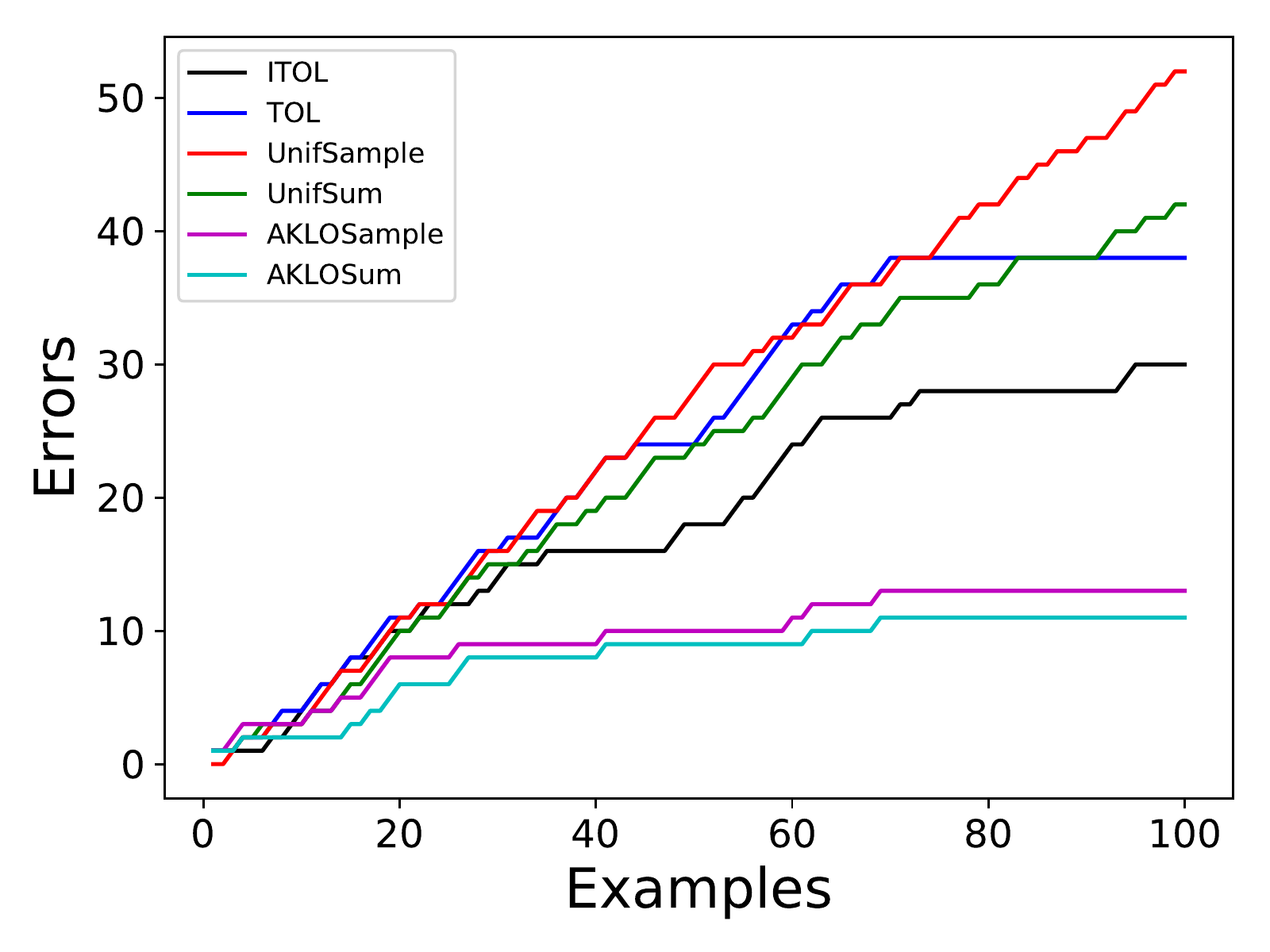}}\\
					
\caption{Evolution of performances of tested algorithms on different datasets: (a)-(e) evolution of Average Cumulative Error rate (ACE in $\%$) over the tasks; (f) cumulative errors for the last task in shoes dataset, for one experiment.}
\label{result_fig}
\end{figure*}

Fig.~\ref{result_fig}(a)-(e) show the the evolution of the average cumulative error rate with the tasks, following the same trends reported in Table~\ref{error_table}. Fig.~\ref{result_fig}(f) reports the cumulative errors of one realization of tested algorithms for the last task of the \emph{shoes} dataset --- comparisons of other datasets are provided in the supplementary material. Such results support the claims that even in the context of non i.i.d. tasks and a relatively small scale interactions, the algorithms proposed are still able to provide a good performance for each task. 

Table~\ref{time_table} shows the average running time for each algorithm, where proposed AKLO\,Sample and AKLO\,Sum algorithms demonstrates their time efficiency. We should also point out that for the \emph{Spam} dataset, the execution time are longer than the others datasets given the extremely high dimensionality of the observations.  Moreover, Unif\,Sample and AKLO\,Sample approaches take a little more running time, due to the sampling operation. 

\begin{table*}[!tbp]
  \centering
  \begin{tabular}{@{}ccccccc@{}}
    \toprule
   \textbf{Dataset}  & \textbf{ITOL}    & \textbf{TOL} & \textbf{Unif\,Sample} & \textbf{Unif\,Sum} & \textbf{AKLO\,Sample} & \textbf{AKLO\,Sum} \\
    \midrule
    \textbf{Syn1} & $41.10 $\tiny$\pm 1.80$ & $22.08 $\tiny$\pm 2.11$ & $40.77 $\tiny$\pm 2.38$ & $35.31$\tiny$\pm 1.87$ &  $13.87 $\tiny$\pm 1.57$ &  $\mathbf{11.00 } $\tiny$\mathbf{\pm 1.52}$ \\
    \textbf{Syn2} & $41.85 $\tiny$\pm 1.35$ & $43.35 $\tiny$\pm 2.95$ & $49.75 $\tiny$\pm 0.92$ & $41.84$\tiny$\pm 1.29$ &  $16.06 $\tiny$\pm 2.22$ &  $\mathbf{12.91} $\tiny$\mathbf{\pm 2.42}$ \\
   \textbf{Landmine} & $18.81 $\tiny$\pm 0.33$ & $14.49 $\tiny$\pm 0.37$ & $21.94 $\tiny$\pm 2.10$ & $11.14$\tiny$\pm 0.91$ &  $13.52 $\tiny$\pm 0.76$ &  $\mathbf{10.39 } $\tiny$\mathbf{\pm 0.38}$ \\
    \textbf{Spam} & $24.53 $\tiny$\pm 1.23$ & $17.47 $\tiny$\pm 0.65$ & $30.02 $\tiny$\pm 2.92$ & $16.68$\tiny$\pm 1.66$ &  $18.20 $\tiny$\pm 1.18$ &  $\mathbf{14.30 } $\tiny$\mathbf{\pm 1.32}$ \\
    \textbf{Shoes} & $31.16 $\tiny$\pm 1.97$ & $29.81 $\tiny$\pm 1.87$ & $43.27 $\tiny$\pm 5.10$ & $31.64$\tiny$\pm 3.99$ &  $26.22 $\tiny$\pm 2.96$ &  $\mathbf{20.74 } $\tiny$\mathbf{\pm 3.24}$ \\          
    \bottomrule
  \end{tabular}
  \caption{Average cumulative error rate ($\%$) $\pm$ standard deviation ($\%$) over $10$ repetitions.}
  \label{error_table}
\end{table*}

\begin{table*}[!tbp]
  \centering
  \begin{tabular}{@{}ccccccc@{}}
    \toprule
    \textbf{Dataset}  & \textbf{ITOL}    & \textbf{TOL} & \textbf{Unif\,Sample} & \textbf{Unif\,Sum} & \textbf{AKLO\,Sample} & \textbf{AKLO\,Sum}\\
    \midrule
     \multirow{2}{*}{\textbf{Syn1}}& $0.029$ & $0.023   $ & $0.426   $ & $0.211  $ &  $0.425  $ &  $0.211 $ \\[-5pt]
      & \tiny$\pm 1.57\times 10^{-4}$& \tiny$\pm 4.39 \times 10^{-4}$ & \tiny$\pm 3.63\times 10^{-4}$ & \tiny$\pm 7.77\times 10^{-4}$ & \tiny $\pm 6.89\times 10^{-4}$ &\tiny  $\pm 5.47\times 10^{-4}$ \\  [4pt]
    \multirow{2}{*}{\textbf{Syn2}} & $0.028  $ & $0.028  $ & $0.418  $ & $0.205 $ &  $0.416  $ &  $0.205  $ \\[-5pt]
      & \tiny$ \pm 1.25\times 10^{-4}$ & \tiny$  \pm 5.42 \times 10^{-4}$ &\tiny $  \pm 4.93\times 10^{-4}$ & \tiny$ \pm 4.00\times 10^{-4}$ & \tiny $  \pm 3.70\times 10^{-4}$ & \tiny $ \pm 3.47\times 10^{-4}$ \\[4pt] 
   \multirow{2}{*}{\textbf{Landmine}} & $0.022 $ & $0.019 $ & $0.361  $ & $0.179 $ &  $0.360  $ &  $0.179  $ \\[-5pt]
      &\tiny $  \pm 7.82\times 10^{-5}$ & \tiny$  \pm 7.92 \times 10^{-5}$ &\tiny $  \pm 2.36\times 10^{-3}$ &\tiny $ \pm 3.30\times 10^{-4}$ &\tiny  $  \pm 1.18\times 10^{-3}$ &  \tiny$  \pm 6.17\times 10^{-4}$ \\[4pt] 
    \multirow{2}{*}{\textbf{Spam}} & $0.380  $ & $0.398 $ & $2.260  $ & $2.249 $ &  $2.467 $ &  $2.238 $ \\[-5pt]
     & \tiny$ \pm 0.012$ &\tiny $ \pm 0.011$ & \tiny$  \pm 0.015$ &\tiny $ \pm 0.019$ & \tiny $  \pm 0.011$ &\tiny  $ \pm 0.011$ \\ [4pt]
   	\multirow{2}{*}{\textbf{Shoes}} & $0.006  $ & $0.006  $ & $0.080  $ & $0.043 $ &  $0.080  $ &  $0.043  $ \\ [-5pt]
   	  & \tiny$  \pm 5.73\times 10^{-5}$ &\tiny$  \pm 6.67 \times 10^{-5}$ & \tiny$ \pm 1.44\times 10^{-4}$ &\tiny $ \pm 1.23\times 10^{-4}$ &\tiny  $  \pm 1.63\times 10^{-4}$ & \tiny $  \pm 1.02\times 10^{-4}$ \\ 
    \bottomrule
  \end{tabular}
  \caption{Average running time (seconds) $\pm$ standard deviation (seconds) over $10$ repetitions.}
  \label{time_table}
\end{table*}

\section{Conclusion}

We are proposing a novel lifelong online learning framework to deal with consecutive online learning tasks relying on knowledge accumulated during past experiences. Specific methods are given to effectively leverage the predictions from the current task classifier and the models built for previous tasks. A theoretical analysis shows the effectiveness of the proposed method even without assumptions on the distributions. Several experiments on both synthetic and real datasets with different contexts are also providing an empirical validation of the proper behaviour of the proposed algorithms.

\bibliography{example_paper} 
\bibliographystyle{aaai}

\end{document}